\documentclass{article}

\usepackage{arxiv}

\usepackage[utf8]{inputenc} 
\usepackage[T1]{fontenc}    
\usepackage{hyperref}       
\usepackage{url}            
\usepackage{booktabs}       
\usepackage{amsfonts}       
\usepackage{nicefrac}       
\usepackage{microtype}      
\usepackage{lipsum}		
\usepackage{graphicx}
\usepackage{natbib}
\usepackage{doi}

\usepackage{graphicx}
\usepackage{multirow}
\usepackage{booktabs}
\usepackage{booktabs,tabularx,makecell}
\newcolumntype{L}[1]{>{\raggedright\arraybackslash}p{#1}}
\newcolumntype{Y}{>{\raggedright\arraybackslash}X}

\usepackage[most]{tcolorbox}

\usepackage{adjustbox}

\title{SynCED-EnDe 2025: A Synthetic and Curated English  - German Dataset for Critical Error Detection in Machine Translation}

\author{
  Muskaan Chopra\textsuperscript{*} \\
  Rheinische Friedrich-Wilhelms-Universität Bonn, Germany \\
  Lamarr Institute for Machine Learning and Artificial Intelligence, Germany \\
  \texttt{chopra.muskaan@uni-bonn.de} \\
  \And
  Lorenz Sparrenberg\textsuperscript{*} \\
  Rheinische Friedrich-Wilhelms-Universität Bonn, Germany \\
  Lamarr Institute for Machine Learning and Artificial Intelligence, Germany \\
  \texttt{sparrenberg@bit.uni-bonn.de} \\
  \And
  Rafet Sifa\textsuperscript{*} \\
  Rheinische Friedrich-Wilhelms-Universität Bonn, Germany \\
  Fraunhofer IAIS, Sankt Augustin, Germany \\
  Lamarr Institute for Machine Learning and Artificial Intelligence, Germany \\
  \texttt{rafet.sifa@iais.fraunhofer.de} \\
}

\date{}

\renewcommand{\shorttitle}{\textit{arXiv} Template}

\begin{document}

\maketitle
\footnotetext[1]{Equal contribution.}

\begin{abstract}
Critical Error Detection (CED) in machine translation aims to determine whether a translation is safe to use or contains unacceptable deviations in meaning. While the WMT21 English-German CED dataset provided the first benchmark, it is limited in scale, label balance, domain coverage, and temporal freshness. We present \textbf{SynCED-EnDe}, a new resource consisting of 1,000 gold-labeled and 8,000 silver-labeled sentence pairs, balanced 50/50 between error and non-error cases. SynCED-EnDe draws from diverse 2024-2025 sources (StackExchange, GOV.UK) and introduces explicit error subclasses, structured trigger flags, and fine-grained auxiliary judgments (obviousness, severity, localization complexity, contextual dependency, adequacy deviation). These enrichments enable systematic analyses of error risk and intricacy beyond binary detection.  
The dataset is permanently hosted on GitHub\footnote{\url{https://github.com/muskaan712/SynCED\_EnDe\_2025}} 
and Hugging Face\footnote{\url{https://huggingface.co/datasets/moon712/SynCED\_EnDe\_2025}}, 
accompanied by documentation, annotation guidelines, and baseline scripts. Benchmark experiments with XLM-R and related encoders show substantial performance gains over WMT21 due to balanced labels and refined annotations. We envision SynCED-EnDe as a community resource to advance safe deployment of MT in information retrieval and conversational assistants, particularly in emerging contexts such as wearable AI devices.
\end{abstract}

\section{Introduction}

Large language models (LLMs) and neural machine translation (MT) systems are 
increasingly embedded in everyday information access scenarios, from wearable 
devices such as Apple AirPods~\cite{airpods3} and Meta Glasses~\cite{metaglasses} 
to mobile assistants and professional communication platforms. In these contexts, 
translation quality is not merely a matter of fluency: while minor stylistic issues may be tolerable, \emph{critical errors}, serious meaning deviations that can cause misunderstanding or harm, remain a pressing concern 
\cite{federmann2021findings,specia2020findings}. The task of 
\emph{Critical Error Detection (CED)} directly addresses this challenge by requiring models to decide whether a translation is safe (\textsc{not}) or contains an error (\textsc{err}). 

The WMT21 shared task on CED for English-German~\cite{federmann2021findings} 
provided the first benchmark dataset for this problem. While highly valuable, it is restricted to Wikipedia comments, collected before 2021, relatively small in scale, and strongly imbalanced, with only $\approx$28\% of translations labeled as errors. Although the task defined categories such as toxicity, negation, numbers, named entities, and safety, these were not consistently annotated in the released resource. The subsequent WMT22 shared task~\cite{federmann2022findings} extended CED to additional tracks, but similar limitations persisted.

To address these issues, we introduce \textbf{SynCED-EnDe} (Synthetic + Curated Error Detection, English-German), a new dataset designed as a community resource for robust and reproducible research in MT evaluation and information retrieval. SynCED-EnDe provides 1,000 gold-labeled evaluation pairs (with manual verification) and 8,000 silver-labeled training pairs, both balanced 50/50 between \textsc{err} and \textsc{not}. Data are drawn from diverse 2024-2025 sources such as StackExchange and GOV.UK guidance, ensuring temporal freshness. Beyond binary labels, the dataset includes explicit error subclasses and structured trigger flags, and the evaluation split is enriched with auxiliary judgments across five dimensions (error obviousness, severity, localization complexity, contextual dependency, and adequacy deviation). Together, SynCED-EnDe and WMT21/22 provide complementary testbeds: the latter reflect noisy human annotation in a narrow domain, while the former offers a balanced, multi-domain, and temporally fresh benchmark for probing critical error detection in MT and IR scenarios. 

\section{Related Work}

Automatic evaluation of MT has traditionally relied on reference-based metrics such as BLEU~\cite{papineni2002bleu} and METEOR~\cite{banerjee2005meteor}, and more recently on neural learned metrics such as COMET~\cite{rei2020comet} and 
COMETKiwi~\cite{rei2022cometkiwi}. While these metrics estimate translation quality, they do not directly address the binary decision of whether a translation is safe to use or contains a critical error. Related to error-specific evaluation, Bauckhage et al.\ \cite{9003090} introduced the first large-scale resource for \emph{contradiction detection in German}, created via machine translation of SNLI. Their work demonstrates the utility of translation-driven dataset construction for semantic validity tasks, aligning with our approach of synthetic error generation in SynCED-EnDe.

This binary framing was introduced in the WMT CED shared tasks 
\cite{federmann2021findings,federmann2022findings,specia2020findings}. 
For English-German, encoder-based models such as XLM-R reached an MCC of 0.46 on 
WMT21~\cite{pucknat2022informed}, establishing a baseline for subsequent work. 
However, both WMT21 and WMT22 remain limited by domain restriction, outdated data, label imbalance, and the absence of auxiliary judgments.

Beyond binary classification, annotation frameworks such as MQM 
\cite{lommel2014multidimensional} and direct assessment corpora 
\cite{graham2017can} capture detailed error categories, while more recent work has explored LLMs for automatic quality judgments 
\cite{kocmi2023lm_eval,lu2023erroranalysis,peng2023towards}. Although these show 
strong correlation with human evaluation, they raise concerns about consistency and pretraining contamination. In parallel, datasets tailored to CED have emerged: Pucknat et al.~\cite{pucknat2022informed} investigated informed pre-training, while Jung et al.~\cite{jung2024explainable} released an explainable CED dataset with error types and explanations across language pairs. These efforts underscore the importance of detecting consequential errors, but remain either limited in scale, costly to extend, or temporally outdated.

From an information retrieval perspective, reliable MT is increasingly vital in 
wearable and conversational AI applications, where mistranslations directly affect retrieval, search, and recommendation quality. In such scenarios, 
\emph{trustworthy translations} are critical for user safety and satisfaction. 
Our work contributes to this line by providing SynCED-EnDe, a balanced, temporally fresh, and systematically enriched dataset that directly addresses these gaps.

\section{Dataset Creation and Annotation}
\label{sec:data}
\begin{figure}[!htbp]
\centering
\includegraphics[width=\linewidth]{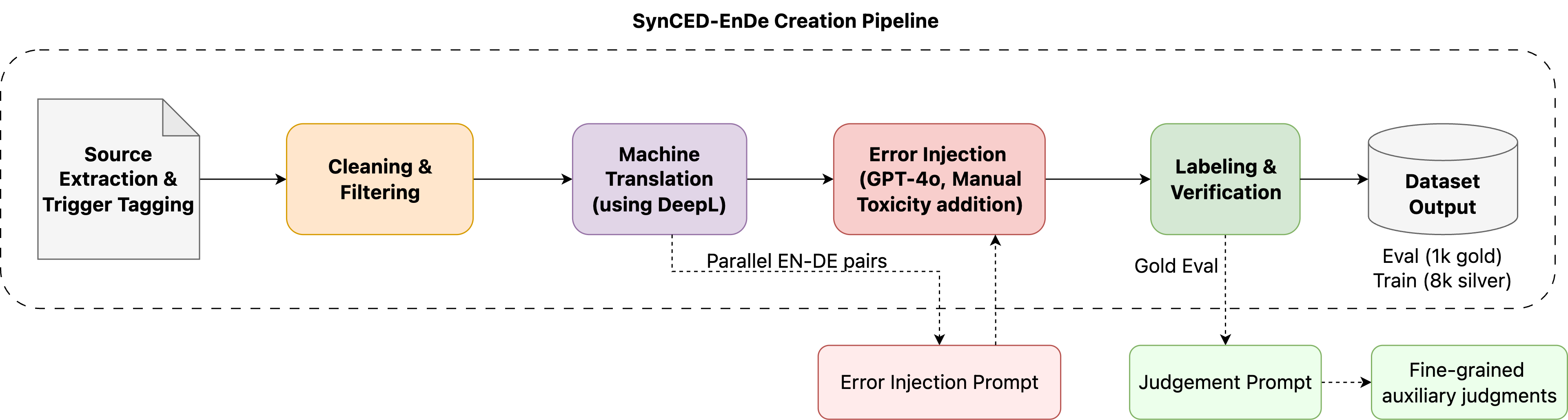}
\caption{SynCED-EnDe pipeline: sources from 2024-2025 (StackExchange, GOV.UK) 
are cleaned and translated (EN$\rightarrow$DE). Controlled errors are injected with prompts 
(lexical, numerical, negation, toxicity), and labels are refined via prompt-based rechecking: 
three LLM rounds for training, three LLM rounds + manual correction for evaluation.}

\label{fig:pipeline}
\end{figure}

The SynCED-EnDe dataset was built through a multi-stage pipeline combining data collection, preprocessing, translation, error injection, and labeling. Figure~\ref{fig:pipeline} provides an overview of the process.

\subsection{Data Collection}
We extracted English source sentences from diverse publicly available domains including Stack Exchange (travel, health, workplace, aviation, DIY, skeptics) and GOV.UK guidance documents. To ensure temporal freshness and reduce overlap with large language model (LLM) pretraining data, only content published between August 2024 and August 2025 was included.  

Preprocessing consisted of sentence cleaning, splitting, and de-duplication, with a maximum length of 30 tokens. During harvesting, four trigger flags were automatically tagged for later analysis: named entities (\texttt{has\_ner}, detected with spaCy \cite{spacy2020}), numerical expressions (\texttt{has\_num}), negation terms (\texttt{has\_neg}), and safety-critical keywords (\texttt{is\_saf\_seed}).  

\subsection{Translation and Error Injection}
The filtered English sentences were translated into German using the commercial MT system DeepL.\footnote{\url{https://www.deepl.com/}} Controlled translation errors were then introduced using GPT-4o.\footnote{\url{https://openai.com/research/gpt-4o}} Categories included lexical substitutions, numerical distortions, and negation flips.  
Because LLMs rarely generate explicitly toxic errors, we created a dedicated toxicity subclass by modifying neutral translations into toxic or unsafe variants (e.g., inserting slurs, offensive adjectives, or unsafe imperatives). This subclass provides additional coverage of safety-critical errors, complementing the safety seed triggers.  
This step was driven by a dedicated injection prompt, which instructed the model to inject one controlled error type per sample while preserving all other content (see Box~\ref{prompt:inject}).

\begin{tcolorbox}[colback=gray!5,colframe=black!50,
                  title=Error Injection Prompt\label{prompt:inject}]
\scriptsize
You are a data generator for critical error detection (CED).  
Given a correct German translation, inject a \emph{controlled} critical error 
according to the requested error type. Only return the rewritten German.\\

Error types:\\
- NAM: corrupt, drop, or replace a named entity.\\
- NUM: change a number, unit, or date/time.\\
- SEN: flip negation or sentiment polarity.\\
- SAF: remove or change a safety/health-critical phrase.\\
- TOX: hallucinate toxicity (mild hate/profanity) not present in the source.\\

Do not explain. Return only the modified German sentence.
\end{tcolorbox}

\subsection{Labeling and Quality Control}
The \textbf{evaluation set} consists of 1,000 pairs (500 \textsc{err}, 500 \textsc{not}). Each example underwent three independent LLM-based rechecks followed by manual verification, ensuring high-quality gold labels. Fine-grained auxiliary judgments (obviousness, severity, localization complexity, contextual dependency, and adequacy deviation) were collected \emph{only for this evaluation set}.  
LLM refinements and auxiliary ratings were obtained using dedicated judgment prompts, which enforced structured outputs and guided consistent use of the 1-5 scales (see Box~\ref{prompt:judge}).

The \textbf{training set} consists of 8,000 pairs (4k \textsc{err}, 4k \textsc{not}), constructed using the same pipeline but verified through three rounds of LLM-based checking without manual validation. No auxiliary judgments are provided for the training split. Overlap between the two splits was prevented by explicit de-duplication.

\begin{tcolorbox}[colback=gray!5,colframe=black!50,
                  title=Judgment Prompt (Auxiliary Ratings)\label{prompt:judge}]
\scriptsize
You are a bilingual annotator judging EN$\rightarrow$DE translation quality for 
critical error detection (CED).

You will receive:
- English source sentence (\texttt{src\_en})
- German translation (\texttt{mt\_de})
- Back-translation of German into English (\texttt{bt\_en})

\textbf{Task:} Rate \emph{five aspects} using a full 1-5 scale. 
Avoid extreme ratings (1 or 5) unless clearly justified. Use 2-4 generously for most cases.  

Return exactly five integers, separated by \textbf{TABS}, in this order:
\texttt{error\_obviousness} \quad
\texttt{error\_severity} \quad
\texttt{localization\_complexity} \quad
\texttt{contextual\_dependency} \quad
\texttt{adequacy\_deviation}

\medskip
\textbf{Aspect definitions:}
\begin{itemize}\setlength\itemsep{1pt}
  \item \textbf{error\_obviousness:} 1 = very obvious, 5 = extremely subtle
  \item \textbf{error\_severity:} 1 = harmless, 5 = critical inversion
  \item \textbf{localization\_complexity:} 1 = one token, 5 = structural scope
  \item \textbf{contextual\_dependency:} 1 = visible in sentence, 5 = requires doc-level expertise
  \item \textbf{adequacy\_deviation:} 1 = minimal drift, 5 = opposite/false meaning
\end{itemize}

\textbf{Output format (strict):} Only five integers separated by tabs.  
Example: \texttt{2\textbackslash t3\textbackslash t2\textbackslash t3\textbackslash t4}
\end{tcolorbox}

\subsection{Output Format}

The dataset is released in tab-separated format with the following fields: unique identifier, English source sentence (\texttt{src\_en}), German translation (\texttt{mt\_de}), source domain and date, trigger flags (\texttt{has\_ner}, \texttt{has\_num}, \texttt{has\_neg}, \texttt{is\_saf\_seed}), binary classification label (\textsc{err} vs.\ \textsc{not}), and optionally an error subclass (lexical, numerical, negation, toxicity, or other).  
  
\section{Dataset Statistics}

\subsection{WMT21 CED (English-German)}
The WMT21 Critical Error Detection (CED) dataset for English-German consists of approximately 8000 training sentence pairs, with 1000 development and 1000 test pairs \cite{federmann2021findings}. Each pair was annotated by three human annotators, with the final binary label (\textsc{err} vs.\ \textsc{not}) determined by majority vote. The data is drawn from Wikipedia comments, and while the task defined critical error categories (toxicity, negation, numbers, named entities, safety), these were not consistently annotated in the released resource. The dataset is also imbalanced, with errors making up only $\approx$28\% of the English-German training set. The subsequent WMT22 shared task \cite{federmann2022findings} extended CED to additional tracks, but many of the same limitations remained, particularly label imbalance and lack of auxiliary dimensions.

\subsection{SynCED-EnDe}
SynCED-EnDe consists of two splits: a gold-labeled evaluation set of 1000 sentences and a silver-labeled training set of 8000 sentences. Both are balanced between \textsc{err} and \textsc{not} classes (Table~\ref{tab:main-stats}).  

The evaluation set underwent three rounds of LLM-based re-annotation followed by manual verification, and includes fine-grained auxiliary judgments (obviousness, severity, localization complexity, contextual dependency, and adequacy deviation). The training set was constructed using the same pipeline but validated through three rounds of LLM-based rechecking only, without manual intervention, and does not include auxiliary judgments. This separation ensures that evaluation remains high-quality and reliable, while training remains cost-efficient.

\setlength{\tabcolsep}{10pt} 
\begin{table}[!htbp]
\centering
\caption{Final released splits of SynCED-EnDe. Both are balanced between \textsc{err} and \textsc{not}.}
\label{tab:main-stats}
\begin{tabular}{lccc}
\toprule
\textbf{Split} & \textbf{Total} & \textbf{\textsc{err}} & \textbf{\textsc{not}} \\
\midrule
Evaluation (gold) & 1,000 & 500 & 500 \\
Training (silver) & 8,000 & 4,000 & 4,000 \\
\bottomrule
\end{tabular}
\end{table}

Source sentences were drawn from StackExchange (workplace, health, travel, aviation, DIY, skeptics) and GOV.UK guidance. Table~\ref{tab:subclass} shows subclass counts for the released splits.

\setlength{\tabcolsep}{10pt}
\begin{table}[!htbp]
\centering
\caption{Distribution of error subclasses in the ERR label in released splits.}
\label{tab:subclass}
\begin{tabular}{lcc}
\toprule
\textbf{Error Subclass} & \textbf{Eval (gold)} & \textbf{Train (silver)} \\
\midrule
NER       & 100 & 800 \\
Numerical & 100 & 800 \\
Negation  & 100 & 800 \\
Toxicity  & 100 & 800 \\
Health    & 100 & 800 \\
\bottomrule
\end{tabular}
\end{table}

\subsection{Comparison with WMT21}
Table~\ref{tab:compare-wmt21} summarizes the key differences between WMT21 and SynCED-EnDe. While WMT21 remains historically important as the first benchmark for critical error detection, SynCED-EnDe introduces balanced splits, explicit subclassing, multiple domains, auxiliary judgments (eval only), and a temporally fresh scope. 

\setlength{\tabcolsep}{10pt}
\begin{table}[!htbp]
\centering
\caption{Comparison of dataset properties between WMT21 CED (English-German) and SynCED-EnDe.}
\label{tab:compare-wmt21}
\begin{tabularx}{\linewidth}{p{3.6cm}X X}
\toprule
\textbf{Property} & \textbf{WMT21 CED (En-De)} & \textbf{SynCED-EnDe} \\
\midrule
Size (train) & $\approx$8000 pairs & 8,000 pairs \\
Size (dev/test) & 1,000 pairs each & 1,000 (eval) \\
Label balance (train) & ERR $\approx$27.9\%, NOT $\approx$72.1\% & ERR = 50\%, NOT = 50\% \\
Label balance (eval) & ERR $\approx$25-30\%, NOT $\approx$70-75\% & ERR = 50\%, NOT = 50\% \\
Domains & Wikipedia comments & StackExchange, GOV.UK \\
Error categories & Defined (TOX, NER, NUM, NEG, SAF) but not consistently annotated & Explicit trigger flags + toxicity subclass \\
Annotation & 3 human annotators, majority vote & LLM recheck + manual (gold), LLM only (silver) \\
Auxiliary judgments & None & Provided for eval only (5 dimensions) \\
Temporal scope & Collected before 2021 & 2024-2025 \\
\bottomrule
\end{tabularx}
\end{table}

These statistics highlight SynCED-EnDe’s commitment to fairness through balanced labeling, to reliability via rigorous validation protocols, and to reproducible research by providing detailed attribute coverage and clear documentation-all priorities central to advancing trustworthy and societally beneficial information retrieval.

\section{Evaluation Dimensions}
\label{sec:eval-dim}

While SynCED-EnDe remains binary-labeled (\textsc{err} vs.\ \textsc{not}) for
compatibility with WMT-style critical error detection, we enrich the
\emph{evaluation set} with fine-grained auxiliary judgments. These judgments
enable systematic analysis of error profiles beyond a single yes/no label and
facilitate more nuanced evaluation protocols. No auxiliary judgments are
provided for the training split.

\subsection{Judgment Protocol}

We adopt a structured annotation scheme using GPT-4o (refer to prompt in section \ref{sec:data}). 
Each English-German pair in the evaluation set is rated on five independent 
$1$-$5$ scales. \emph{Error obviousness} measures how easy the error is to spot  ($1=$ very obvious, $5=$ extremely subtle). \emph{Error severity} captures the  potential real-world harm if the error goes unnoticed ($1=$ harmless, $5=$ critical inversion). \emph{Localization complexity} reflects how widely the error is spread ($1=$ token-level, $5=$ distributed or structural). \emph{Contextual dependency} indicates the extent of background knowledge required to recognize the error ($1=$ self-contained, $5=$ expert or document-level). Finally, \emph{adequacy deviation}  measures how far the meaning drifts from the source ($1=$ stylistic shift, $5=$ opposite meaning). 
This protocol follows earlier work on direct assessment \cite{graham2017can} and Multidimensional Quality Metrics (MQM) \cite{lommel2014multidimensional}, but focuses on dimensions most relevant for safety-critical error detection and IR applications.

\subsection{Derived Scores}

From the five auxiliary dimensions we compute two composite metrics: the 
\emph{Risk Score} and the \emph{Intricacy Score}. For conciseness, we denote 
severity as $S$, adequacy deviation as $A$, obviousness as $O$, localization 
complexity as $L$, and contextual dependency as $C$.  

The Risk Score emphasizes the potential consequences if an error is missed and 
is defined as
\begin{equation}
\text{Risk} = 0.6S + 0.4A.
\end{equation}
Here, severity receives slightly higher weight since real-world harm is 
primarily determined by the seriousness of an error, while adequacy captures 
the semantic drift from the source.  

The Intricacy Score emphasizes how difficult an error is to detect and is 
defined as
\begin{equation}
\text{Intricacy} = 0.35(6-O) + 0.20L + 0.20C + 0.25A.
\end{equation}
The inverse of obviousness is weighted most heavily, as subtle errors are the 
hardest to identify. Localization and context reflect the dispersion of the 
error and the external knowledge needed, while adequacy again accounts for 
semantic deviation.  

Together, Risk and Intricacy characterize both \emph{dangerous} and 
\emph{hard-to-spot} errors, enabling downstream applications such as 
risk-aware retrieval, re-ranking, and calibration of conversational assistants.

\subsection{Distributional Comparison}

\begin{figure}[!htbp]
  \centering
  \includegraphics[width=\linewidth]{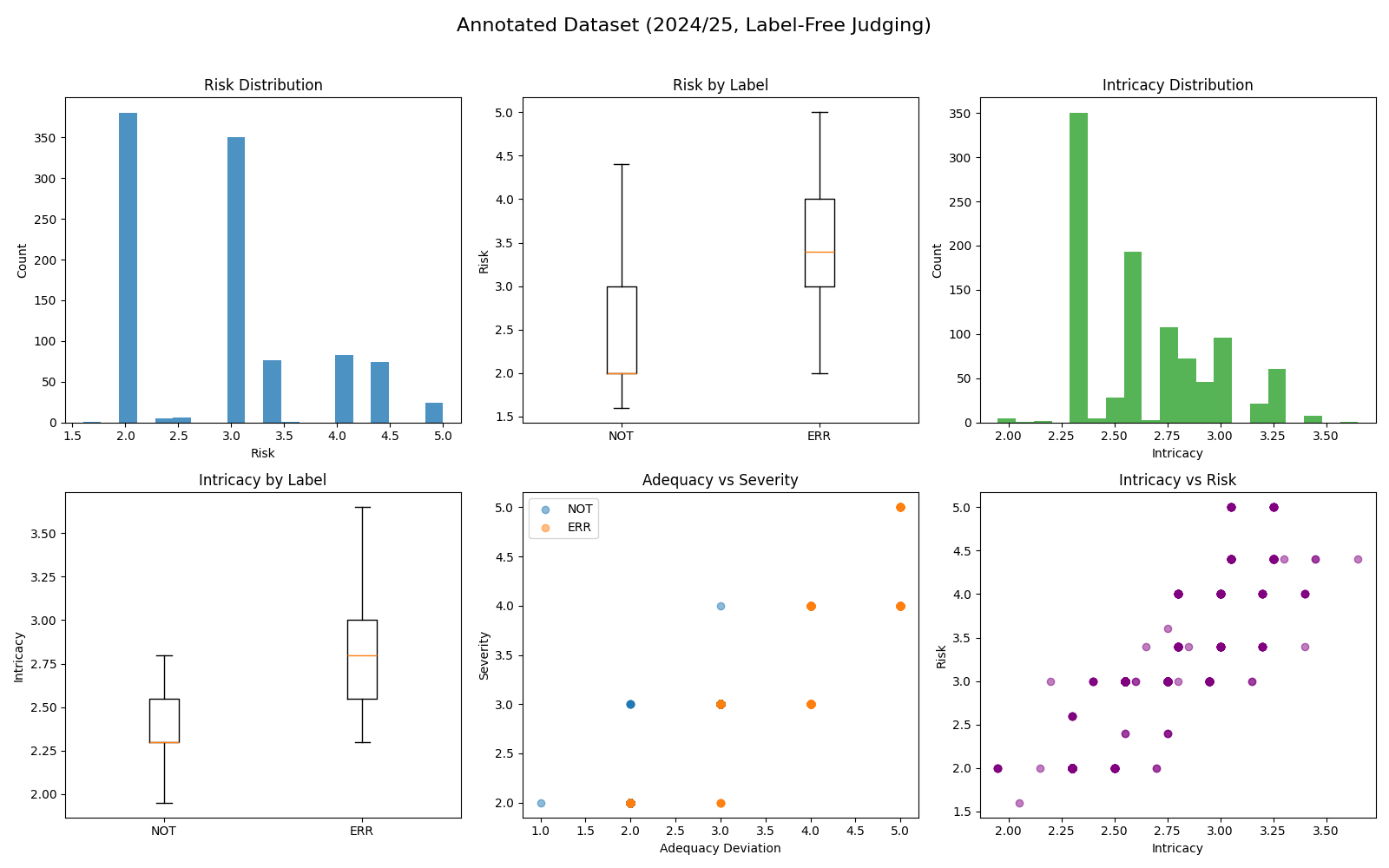}
  \caption{Evaluation dimensions in SynCED-EnDe (evaluation set). Balanced
  risk and intricacy distributions with meaningful separation of
  \textsc{err}/\textsc{not}, covering a wide range of subtle and severe cases.}
  \label{fig:annotated_subplots}
\end{figure}

\begin{figure}[!htbp]
  \centering
  \includegraphics[width=\linewidth]{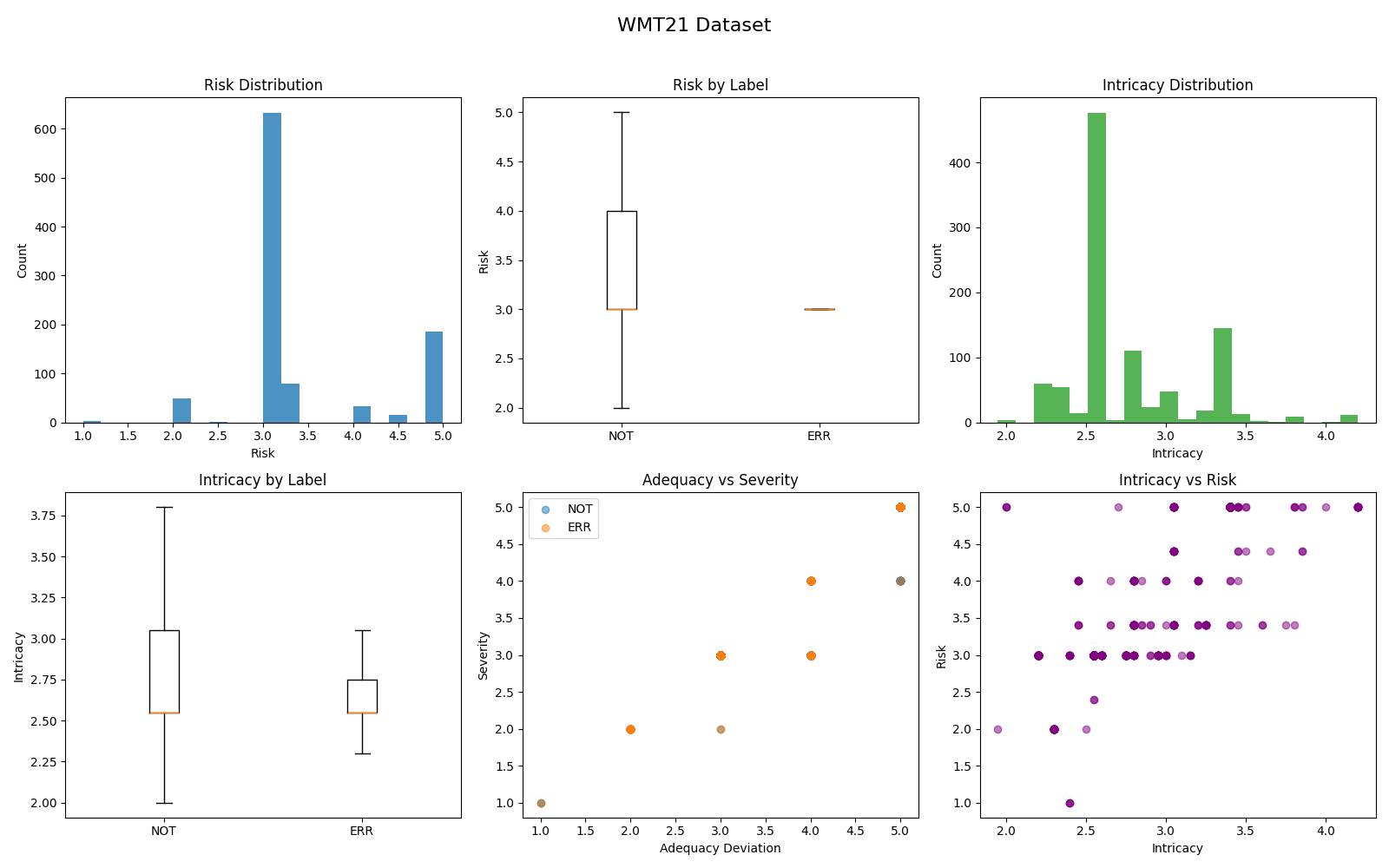}
  \caption{Evaluation dimensions in WMT21. Skewed distributions with many
  obvious or extreme cases, and weaker separation between \textsc{err} and
  \textsc{not}.}
  \label{fig:wmt21_subplots}
\end{figure}

Figures~\ref{fig:annotated_subplots} and~\ref{fig:wmt21_subplots} compare
evaluation-dimension distributions for SynCED-EnDe and WMT21. Our dataset
achieves balanced coverage across risk and intricacy ranges, avoids sharp
peaks, and shows clear separation between \textsc{err} and \textsc{not}. By
contrast, WMT21 is skewed towards extremes, with many obvious or trivial cases
and fewer subtle, moderate-risk errors.

\subsection{Correlation Analysis}

\begin{figure}[!htbp]
  \centering
  \includegraphics[width=0.9\linewidth]{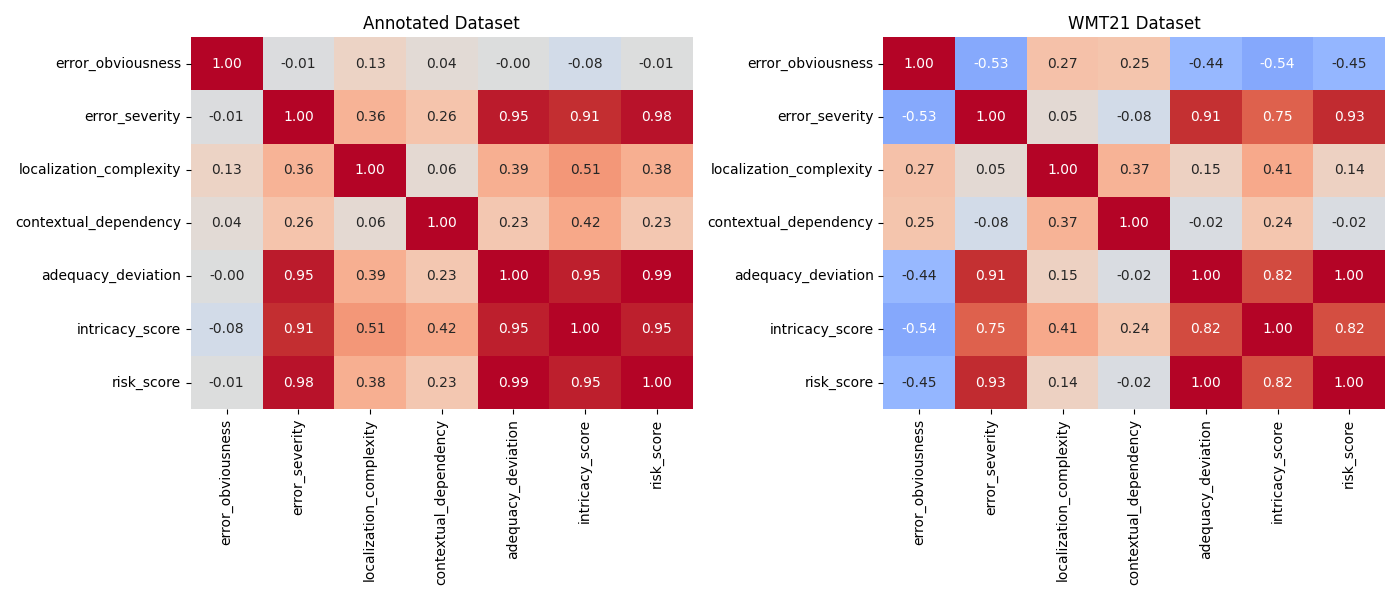}
  \caption{Correlation heatmaps of evaluation dimensions. Left: SynCED-EnDe
  shows consistent and interpretable correlations between adequacy, severity,
  risk, and intricacy, with obviousness largely independent. Right: WMT21
  shows mixed or negative correlations, indicating less consistent annotation
  patterns.}
  \label{fig:correlations}
\end{figure}

Beyond distributions, we also examine how dimensions co-vary
(Figure~\ref{fig:correlations}). For SynCED-EnDe, adequacy deviation,
severity, risk, and intricacy are strongly correlated, while obviousness is
largely orthogonal; a realistic axis where subtle errors can still be highly
consequential. 
By contrast, WMT21 shows inconsistent and sometimes negative
correlations, suggesting annotation noise or mismatched criteria across
annotators.

\newpage
\section{Baseline Results}
\label{sec:baselines}

To illustrate the usability of SynCED-EnDe as a benchmark resource, we report 
baseline results with standard encoder-only architectures. Following prior work 
on CED \cite{federmann2021findings,specia2020findings}, we adopt XLM-R 
\cite{conneau2020unsupervised} as a reference point. On the WMT21 English-German 
CED dataset, XLM-R achieves a Matthews correlation coefficient (MCC) of 0.46, 
with $F_{1}$-ERR = 0.599 and $F_{1}$-NOT = 0.855. These scores have been widely 
cited as the baseline benchmark for binary critical error detection. 

For SynCED-EnDe, we train the same architecture on the 8000 silver-labeled 
training split and evaluate on the 1000 gold evaluation split. Balanced labels 
and temporally fresher content substantially improve learnability: XLM-R reaches 
MCC = 0.819, $F_{1}$-ERR = 0.894, and $F_{1}$-NOT = 0.911. These results show 
that SynCED-EnDe offers a stable and reliable benchmark while remaining directly 
comparable to WMT21.

To contextualize these baselines, we also report approximate parameter counts for each encoder variant (Table~\ref{tab:baselines_enc}). This makes clear that the strong gains observed on SynCED-EnDe are not merely a function of scaling, but hold consistently across both smaller (110M) and larger (560M) architectures.

\begin{table*}[!htbp]
\centering
\caption{Encoder-only baselines on WMT21 and SynCED-EnDe (En-De). 
SynCED-EnDe yields higher scores due to balanced labels and refined annotations. 
Best results per dataset are shown in bold.}
\resizebox{\textwidth}{!}{%
\begin{tabular}{l r cc}
\toprule
Model & Params & WMT21 & SynCED-EnDe \\
 & & {\scriptsize (MCC / $F_{1}$-ERR / $F_{1}$-NOT)} & {\scriptsize (MCC / $F_{1}$-ERR / $F_{1}$-NOT)} \\
\midrule
BERT-base-uncased \cite{bert} & 110M & 0.38 / 0.53 / 0.83 & 0.72 / 0.83 / 0.86 \\
ModernBERT-base \cite{modernbert} & 150M & 0.37 / \textbf{0.56} / 0.79 & 0.49 / 0.77 / 0.69 \\
ModernBERT-large \cite{modernbert} & 395M & 0.38 / 0.53 / 0.83 & 0.52 / 0.78 / 0.72 \\
mmBERT-base \cite{mmbert} & 307M & 0.38 / 0.53 / 0.82 & 0.82 / 0.90 / 0.91 \\
XLM-R (large) \cite{conneau2020unsupervised} & 561M & \textbf{0.46} / 0.59 / \textbf{0.85} & \textbf{0.84} / \textbf{0.91} / \textbf{0.92} \\
\bottomrule
\end{tabular}}
\label{tab:baselines_enc}
\end{table*}

These baselines should not be interpreted as evidence that SynCED-EnDe is 
trivial. Rather, the relatively high scores reflect the dataset’s internal 
consistency and balanced label distribution, which allow models to learn 
effectively without being confounded by skew or noise. This makes 
SynCED-EnDe an ideal \emph{controlled benchmark}: it complements noisier, 
human-annotated resources such as WMT21 by providing a cleaner setting in 
which to probe model sensitivity to subtle but consequential errors. Balanced and well-controlled benchmarks give fairer performance estimation, avoiding bias from label imbalance and annotation noise.

Beyond binary classification, SynCED-EnDe also enables richer evaluation: 
the auxiliary dimensions (risk and intricacy) provide signals that can be used 
to calibrate model confidence or to design risk-aware evaluation protocols. 
We therefore recommend that SynCED-EnDe be used alongside WMT21/22 in 
evaluation suites, where the combination of noisy human annotations and 
controlled synthetic annotations offers a more comprehensive benchmark for 
critical error detection in machine translation and related IR tasks.

\section{Discussion}
\label{sec:discussion}

\subsection{Strengths}

SynCED-EnDe introduces several properties that distinguish it from existing
critical error detection datasets. First, both the training and evaluation
splits are balanced 50/50 between \textsc{err} and \textsc{not}, ensuring that
minority-class performance is not confounded by skew. Second, the dataset has a
fresh temporal scope: all content was harvested from 2024-2025 sources such as
StackExchange and GOV.UK, thereby minimizing overlap with current LLM pretraining
corpora and increasing its robustness as a forward-looking benchmark. Third,
beyond binary labels, SynCED-EnDe introduces explicit subclassing and trigger
flags, including a curated toxicity subclass, which enables targeted analyses of
safety-critical errors. Fourth, the gold evaluation split is enriched with
auxiliary dimensions: obviousness, severity, localization, contextual
dependency, and adequacy, which allow the computation of \emph{risk} and
\emph{intricacy} scores, thereby supporting more nuanced evaluation protocols
than binary accuracy or $F_{1}$. Finally, the dataset maintains annotation
consistency: both splits underwent three independent rounds of LLM-based
reannotation, with the evaluation split further refined by manual verification.
This procedure keeps label noise low while remaining cost-effective.

\subsection{Limitations}

At the same time, the dataset has several limitations that must be considered.
Because error injection relies on GPT-4o, some artifacts may not be fully
representative of naturally occurring MT errors. Moreover, although the 8000
training set underwent three rounds of automated rechecking, it was not manually
validated, unlike the 1000 evaluation split. Finally, the current release is
restricted to English-German; extending the resource to additional language
pairs is an important direction for future work.

\subsection{Complementarity}

SynCED-EnDe is not intended to replace WMT resources but rather to complement
them. WMT21 and WMT22 provide noisy, human-annotated data that capture real-world
variability, whereas SynCED-EnDe offers a balanced, temporally fresh, and
systematically enriched benchmark. Used together, these resources cover both
uncontrolled naturalistic error distributions and controlled evaluation
settings. This combination enables more comprehensive benchmarking and
stress-testing of encoder-based baselines as well as emerging LLM systems.

\section{Conclusion and Future Work}
\label{sec:conclusion}

We introduced \textbf{SynCED-EnDe}, a new dataset for critical error detection
in English-German translation. Unlike prior benchmarks, SynCED-EnDe combines
binary labels with auxiliary judgments, balanced error coverage, and content
from 2024-2025. The resource provides a \emph{gold} evaluation split 
(1000, three LLM rechecks plus manual verification) and a \emph{silver} training
split (8000, three LLM rechecks), enriched with toxicity subclasses and trigger
flags. 

Our analyses show that SynCED-EnDe yields high-quality, internally consistent
annotations with balanced risk and intricacy distributions. Baseline results
demonstrate that the dataset is both learnable and stable, complementing
noisier resources such as WMT21/22. Together, these datasets enable both
controlled and naturalistic evaluation of MT error detection.

Looking ahead, we plan to extend SynCED to additional language pairs, broaden
its domain coverage, and incorporate further subclasses such as stylistic or
cultural errors. We also aim to explore semi-automatic pipelines that combine
LLM-based judgments with targeted human validation. SynCED-EnDe will be
released publicly (TSV format with metadata, Hugging Face and GitHub
repositories) to support reproducibility in MT evaluation, safety-critical NLP,
and LLM calibration.

\section*{Acknowledgement}
This research has been partially funded by the Federal Ministry of Education and
Research of Germany and the state of North-Rhine Westphalia as part of the
Lamarr-Institute for Machine Learning and Artificial Intelligence.

\bibliographystyle{unsrtnat}
\bibliography{references}  

\end{document}